\documentclass[letterpaper, 10 pt, conference]{ieeeconf}  

\usepackage{epstopdf}

\usepackage{amsmath,bm}
\usepackage{amsfonts} 
\usepackage{color}
\usepackage{makecell}
\usepackage{centernot}
\usepackage{graphicx} 

\usepackage{subfig}
\usepackage{multirow}
\usepackage{booktabs}

\newtheorem{lemma}{Lemma}
\newtheorem{definition}{Definition}
\newtheorem{theorem}{Theorem}

\usepackage[hidelinks]{hyperref}
\IEEEoverridecommandlockouts                              
\title{\LARGE \bf
Sensor-based Multi-agent Coverage Control with Spatial Separation in Unstructured Environments
}

\author{ Xinyi Wang$^{1}$, Jiwen Xu$^{1}$, Chuanxiang Gao$^{1}$, Yizhou Chen$^{1}$, Jihan Zhang$^{1}$, \\ Chenggang Wang$^{2}$, Yulong Ding$^{3}$, Ben M. Chen$^{1}$ 
\thanks{The work was supported in part by the Research Grants Council of Hong Kong SAR (14206821 and 14217922),  in part by the Hong Kong Centre for Logistics Robotics, and in part by the Young Scientists Fund of the National Natural Science Foundation of China (Grant No. 62303354 and 62303316).}
\thanks{$^{1}$Xinyi Wang, Jiwen Xu, Yizhou Chen, Chuanxiang Gao, Jihan Zhang, and Ben M. Chen are with the Department of Mechanical and Automation Engineering, The Chinese University of Hong Kong, Shatin, N.T., Hong Kong, China (e-mail: xywangmae@link.cuhk.edu.hk;xujiwen@link.cuhk.edu.hk; cxgao@mae.cuhk.edu.hk; yzchen@mae.cuhk.edu.hk;  jhzhang@mae.cuhk.edu.hk; bmchen@cuhk.edu.hk);}
\thanks{$^{2}$Chenggang Wang is with the Department of Automation, and Key Laboratory of System Control and Information Processing, Ministry of Education of China, Shanghai Jiao Tong University, Shanghai 200240, P. R. China. (cgwang-auv@sjtu.edu.cn).}%
\thanks{$^{3}$Yulong Ding is with the Department of Control Science and Engineering, Tongji University, Shanghai, China, and with the Frontiers Science Center for Intelligent Autonomous Systems, Ministry of Education, China; (e-mail: dingyulong@tongji.edu.cn);}
}

\begin{document}

\maketitle
\thispagestyle{empty}
\pagestyle{empty}

\begin{abstract}
  Multi-robot systems have increasingly become instrumental in tackling coverage problems. However, the challenge of optimizing task efficiency without compromising task success still persists, particularly in expansive, unstructured scenarios with dense obstacles. This paper presents an innovative, decentralized Voronoi-based coverage control approach to reactively navigate these complexities while guaranteeing safety. This approach leverages the active sensing capabilities of multi-robot systems to supplement GIS (Geographic Information System), offering a more comprehensive and real-time understanding of environments like post-disaster. Based on point cloud data, which is inherently non-convex and unstructured, this method efficiently generates collision-free Voronoi regions using only local sensing information through spatial decomposition and spherical mirroring techniques. Then, deadlock-aware guided map integrated with a gradient-optimized, centroid Voronoi-based coverage control policy, is constructed to improve efficiency by avoiding exhaustive searches and local sensing pitfalls. The effectiveness of our algorithm has been validated through extensive numerical simulations in high-fidelity environments, demonstrating significant improvements in task success rate, coverage ratio, and task execution time compared with others.

\end{abstract}

\section{Introduction}

Multi-robot systems are extensively employed in coverage tasks across diverse applications, including gas leak detection, pollution source identification, and search and rescue operations \cite{huang2018coverage,wang2022minimum}. 
These systems can efficiently locate Targets of Interest (ToI) in unknown and unstructured environments through interactive data collection. 
Building on their capabilities, integration with Geographic Information Systems (GIS) offers new opportunities for improving efficiency and accuracy of such operations \cite{zhang2022sim}.
\begin{figure}[thpb]
  \centering
  \setlength{\belowcaptionskip}{-0.5cm}
\setlength{\abovecaptionskip}{0.2cm}
  \includegraphics[width =0.7\hsize]{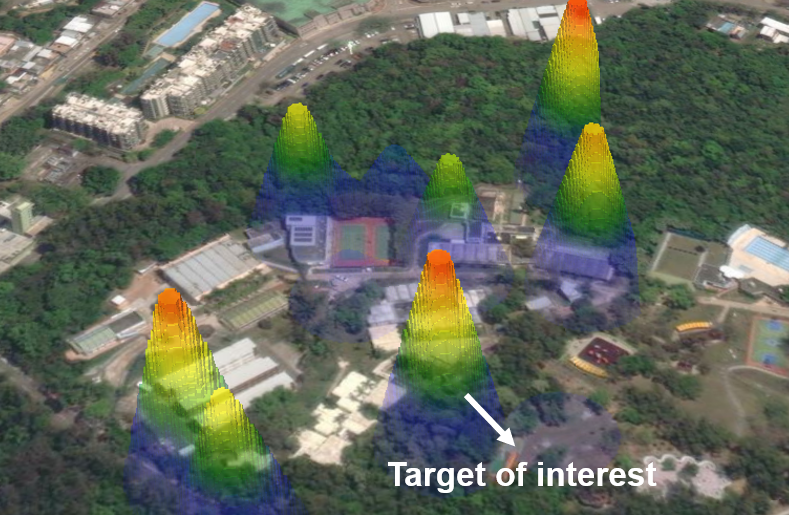}
  \caption{GIS provides maps as prior information for multi-robot systems in post-disaster search and rescue.}
  \label{fig:village}
\end{figure}
While GIS offers extensive pre-collected data, multi-robot information sharing, and advanced analysis features like heat maps as shown in Fig.~\ref{fig:village}, its limitations are evident in uncertain situations such as post-disaster rescue efforts.
In these contexts, the active sensing capabilities of multi-robot systems can serve as real-time supplements to GIS, providing a more comprehensive and up-to-date understanding of the environment.

Despite the unprecedented advancements in this field, the challenge of searching ToI in unknown environments persists, particularly in balancing efficiency with successful task execution.
Some search-based approaches resort to exhaustive exploration of the surrounding environment to identify feasible paths
\cite{lan2016rapidly}.
However, this approach compromises efficiency and necessitates complete environmental traversal. Such inefficiency has critical implications for timely hazard avoidance or life-saving actions, especially given the battery limitations in large-scale environments.

On the other hand, some reactive approaches compute search policies that focus on the immediate rewards associated with ToI, thereby enhancing efficiency through the local sensing range.
These methods, rigorously examined in the context of Centroidal Voronoi Tessellation (CVT) in multi-robot coverage problems \cite{wang2022minimum,luo2019voronoi,mahboubi2012distributed, wang2022time} offer numerous advantages, including asynchronous operation, distributed execution, adaptability, and verifiable asymptotic correctness \cite{cortes2004coverage}. 
However, they are susceptible to local minima due to short-sighted decisions, which can lead to the failure to complete the task.
Furthermore, the majority of existing methods are confined to two-dimensional, well-structured environments \cite{bai2022adaptive}. They fall short in accounting for unstructured settings, including raw sensor data, where robots are often trapped in the presence of non-convex obstacles when attempting to expedite their greedy coverage policy \cite{pierson2017distributed,abdulghafoor2021distributed,wang2023fast}.

In this paper, we introduce a novel decentralized Voronoi-based active coverage policy for unknown and unstructured environments with guaranteed safety. 
This research is driven by scenarios where robots aim to balance coverage efficiency and success rate while adhering to safety constraints.
We aim to develop a cooperative control policy that enables individual robots to make intelligent decisions while enhancing overall team efficiency.
\textbf{Firstly},
we utilize realistic point cloud data as input, characterized by its highly non-convex and unstructured nature. 
By integrating spatial decomposition and spherical mirroring techniques, collision-free Voronoi regions are efficiently generated using only local sensing and position information from other robots.
\textbf{Secondly},
explored information is employed to construct a deadlock-aware guided map to refine subsequent processes for optimal decision-making. This map is integrated with centroid Voronoi based coverage control policy to improve efficiency by avoiding exhaustive searches and local sensing pitfalls.
\textbf{Thirdly},
the effectiveness of our proposed method has been extensively validated through numerical simulations in high-fidelity large-scale environments, including cluttered, thin-structural, and narrow settings. Compared with other Voronoi-based methods, the approach has significantly improved the successful coverage ratio and time.

To the best of our knowledge, we are the first to propose a Voronoi-based coverage algorithm that mitigates the issue of local minima trapping in general unstructured obstacle environments supported by convergence guarantees.


\section{Related Work}

This work integrates contributions related to its functionality, building upon collision avoidance methods and our previous research on coverage control \cite{wang2023fast}.

A crucial consideration in collision avoidance is the formulation of obstacles derived from point cloud data. 
Most existing methods rely on using high resources to maintain a cost map, such as the Euclidean Distance Transform (EDT) map, to calculate the minimum distance to obstacles for optimization \cite{wang2021decentralized,xi2021gto,chen2022gpu}, which may suffer from time-consuming computations and redundant processes \cite{zhou2020ego}.
Alternatively, some strategies depend on down-sampling and spatial partitioning using convex clusters, \cite{liu2017planning}. Unfortunately, such techniques may sacrifice the free space or lead to unsafe areas \cite{toumieh2022decentralized}.
Instead, our approach directly generates safe Voronoi cells on a voxel-based environment representation, inspired by the concept presented in \cite{katz2005mesh,zhong2020generating}. 
The advantages lie in threefold:
1) It operates exclusively on point cloud data, eliminating the need for a post-processed map and ensuring the preservation of raw sensor measurement fidelity;
2) It does not assume the convexity or structure of the environment, making it particularly suitable for real-world sensors influenced by noise and adaptable to a range of scenarios;
3) The strategy requires only the sensing of other robots' positions, which simplifies communication requirements and bolsters system robustness.


Another challenge in active coverage control design arises from the complexities of raw sensor measurements and unstructured environments, which often lead robots to be trapped in local minima. 
This issue is commonly encountered in coverage methods that rely on CVT, a form of non-convex optimization \cite{pierson2017distributed,abdulghafoor2021distributed,wang2023fast}. While there are some techniques like random restarts \cite{oleynikova2016continuous} and iterative refinement \cite{gao2020teach} address this issue, it has not been fully solved.
In this work, we introduce a deadlock-aware guided map that incorporates a move-to-centroid policy and dynamically updates feasible directions as new obstacles are encountered. 
The gradient of this map effectively guides the robot away from obstacles and pitfalls while simultaneously directing it toward the target, thereby preventing entrapment in local minima.


\section{Preliminaries}
Consider a team of $n$ robots localized at $\mathbf{P}(t) = [\mathbf{p}_1(t),\dots,\mathbf{p}_n(t)]$ at time $t$ in a workspace $\mathcal{W} \subset  \mathbb{R}^3$. 
The dynamics of each robot can be modeled as a single integrator, described by $\dot{\mathbf{p}}_i = \mathbf{u}_i$, and its geometry can be modeled as a ball with centered position $\mathbf{p}_i = (p_{i,x},p_{i,y},p_{i,z} )$ and radius $r_i$, where $i \in 1,\dots,n$. 
The environment is represented by a grid map with discretized points $\mathbf{p} \in \mathcal{W}$. 
It can be uniformly partitioned into non-overlapping regions corresponding to each robot utilizing Voronoi partitioning.
These partitions are generated based on $\mathbf{P}$ and can be regarded as the intersection of a set of maximum margin separating hyper-planes \cite{Arslan2019}. For each robot $i$, the Voronoi cell is defined as follows: 
  

 \begin{equation}
\begin{aligned}
\label{voronoi}
\mathcal{V}_i = &\{\mathbf{p} \in \mathcal{W} | \Vert \mathbf{p}-\mathbf{p}_i^*\Vert \leq  \Vert \mathbf{p}-\mathbf{p}_j^*\Vert\text{,} \\
& (\mathbf{p}_i^*, \mathbf{p}_j^*) = \underset{\mathbf{p}_i \in \mathcal{R}_i, \mathbf{p}_j \in \mathcal{R}_j}{\arg\min} d(\mathbf{p}_i, \mathbf{p}_j) \}
\end{aligned}
\end{equation}
where $\Vert \cdot \Vert$ denotes $l_2$-norm.  It can be observed that the discrete points within the Voronoi cell $\mathcal{V}_i$ are closer to its generator $\mathbf{p}_i$ than to any other robots.
Therefore, it is reasonable to assume that each robot $i$ is tasked with its corresponding region $\mathcal{V}_i$ using a limited-range sensor.

 \subsection{Sensor and Obstacle Representation}

 Let $\mathcal{O} = \{\mathbf{q}_o\in \mathcal{W} | \mathbf{q}_o=(q_{o,x},q_{o,y},q_{o,z}), 1 \leq o\leq m \}$ be the set of obstacles, where $m$ is the number of all point clouds.
 The free configuration space is defined as $\mathcal{W}_{\mathrm{free}} =\mathcal{W} \backslash \mathcal{O}.$
 Suppose the robot's perception is solely sensor-based, with the sensor equipped having a fixed sensing range denoted as $R_\mathrm{sensor} \in \mathbb{R}_{>0}$. This implies that the robot's knowledge about the location of environmental obstacles is restricted to the structure it perceives within a nearby area around its current position. Let the robot be surrounded by $m_r$ point clouds of obstacles inside the sensor range $R_\mathrm{sensor}$.
 The observable obstacles environment, represented by a series of discrete points and denoted as the set $\mathcal{Q} = \{ \mathbf{q}_o \in \mathcal{W} \mid \Vert \mathbf{q}_o - \mathbf{p}_i \Vert \leq R_\mathrm{sensor}, 1\leq o\leq m_r \}$, constitutes a spherical region centered at the robot's position $\mathbf{p}_i$.  Any region outside the robot's sensory footprint is assumed to be obstacle-free until updated when the robot perceives it.



\subsection{Voronoi-based Coverage Control}
To evaluate the coverage of the target of interest (ToI) by a multi-robot system, two key factors must be considered: the value of each sensed point and the sensing capability, which is influenced by the distance between the robot and the point on the grid to be sensed.
The value of these points is indicative of the importance of the information to be covered. To quantify this, we introduce a density function  $\phi(\cdot):\mathcal{W} \rightarrow \mathbb{R}^+$ based on a Gaussian mixture model for reference coverage information.
As the distance between the robot and the point to be sensed increases, the sensing capability generally diminishes. To optimize the deployment locations of the multi-robot system, we utilize a cost function based on $\phi(\mathbf{p})$ \cite{cortes2004coverage,abdulghafoor2021distributed,breitenmoser2010voronoi}
\begin{equation}
\begin{aligned}
\label{eq: cost}
\mathcal{H}(\mathbf{P}) = \sum_{i=1}^{n} \mathcal{H}_i(\mathbf{P}) =\sum_{i=1}^{n}\int_{\mathcal{V}_i}\Vert \mathbf{p}-\mathbf{p}_i \Vert^2 \phi(\mathbf{p})d\mathbf{p}.
\end{aligned} 
\end{equation}
The derivative of $\mathcal{H}$ indicates that moving to the centroids of the sensors' corresponding Voronoi cells can reduce the coverage cost. Hence, a gradient decent control policy $\mathbf{u}_i$ is designed to steer robots to a CVT: 
\begin{equation}
\begin{aligned}
\label{eq: u}
\mathbf{u}_i = -u_{\max}\frac{\mathbf{p}_i - \mathbf{C_{\mathcal{V}_i}}}{\Vert \mathbf{p}_i - \mathbf{C_{\mathcal{V}_i}} \Vert}.
\end{aligned} 
\end{equation}
where $\mathbf{C}_{\mathcal{V}_i} = \underset{\mathbf{p}_i}{\arg \min} \mathcal{H}_{i}(\mathbf{P}).$



\section{
Sensor-based Coverage Control with Deadlocks and Collision Avoidance} 


 \begin{figure*}[t]
  \centering
  \setlength{\belowcaptionskip}{-0.6cm}
\setlength{\abovecaptionskip}{0.3cm}
  \includegraphics[width =0.8\hsize]{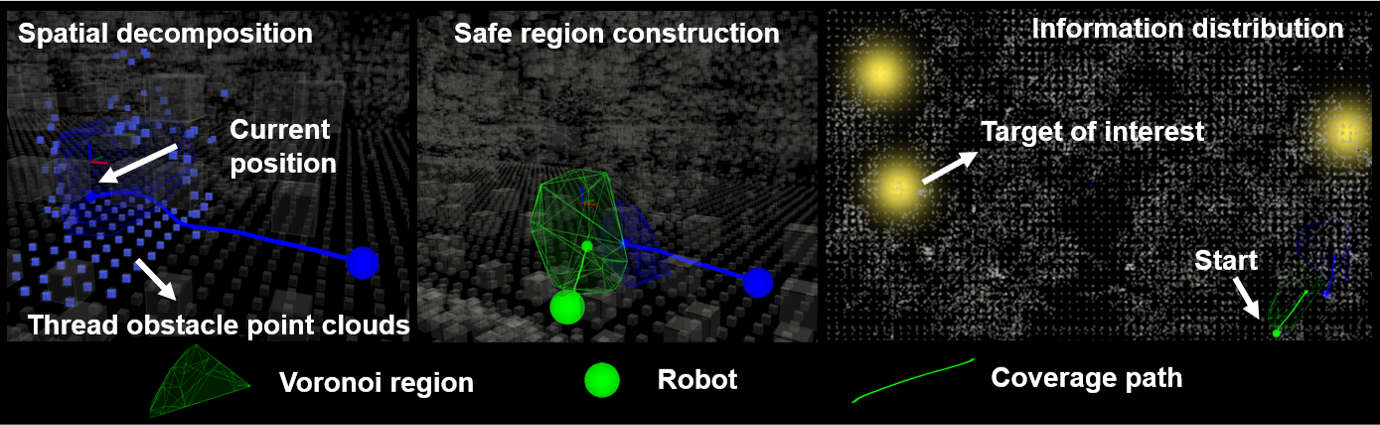}
  \caption{Multi-robot coverage mission in unstructured environments with three sparsely distributed information sources.}
  \label{fig:overview}
\end{figure*}

This section introduces a novel reactive coverage method that enables a multi-robot team to cover the ToI using only its current position and the structure perceived within its local sensor footprint. 
A method overview is depicted in Fig.~\ref{fig:overview}.




\subsection{ Spatial Decomposition}
To separate the movement space of multi-robots into some security domains without executing duplicated tasks, a safe convex region needs to be constructed for each robot at discrete time $k\Delta t$, where $\Delta t$ is the time interval. 
For each time, the Voronoi cell for each robot is only determined by neighboring robots and obstacles, and can thus be formed as the intersection of half-spaces separating robot $i$ from obstacles or other robots with linear separator  $\mathbf{a}_{ij}$, $b_{ij}$, and $\mathbf{a}_{io}$,$b_{io}$, where $i,j \in 1,\dots,n$ and $o \in 1,\dots,m$.

\subsubsection{Robot-robot Collision Avoidance Hyperplane}
To achieve a spatially load-balanced deployment that can provide locally optimal sensor coverage of a target, it is essential to define a separation strategy that ensures that each voxel in the deployment region is closer to the position of the robot, which serves as the generator for its Voronoi cell, than to any other robot in the environment. This separation condition is fundamental to the definition of the Voronoi cell, which partitions the environment into non-overlapping regions based on each robot's location.
We can calculate $\mathbf{a}_{ij}$ and $b_{ij}$ by finding a perpendicular bisector of any two positions of robot $\mathbf{p}_{i},\mathbf{p}_{j}$ using the definition in the following:
\begin{equation}
\begin{aligned}
\label{eq: defvoronoi}
\mathbf{a}_{i j} = \mathbf{p}_{i j}=\mathbf{p}_{i }-\mathbf{p}_{j}, \quad b_{i j} = \mathbf{p}_{i j}^\mathrm{T}\frac{\mathbf{p}_{i}+\mathbf{p}_{j}}{2}.
\end{aligned}
\end{equation}

\subsubsection{Robot-obstacle Collision Avoidance Hyperplane}

The robot operates within a realistic world that contains unstructured obstacles, represented by point cloud data.
These raw data are processed with computationally demanding techniques like surface reconstruction and the maintenance of additional cost maps, commonly used in traditional methods. 
To develop a more efficient and lightweight algorithm, we have devised a method that directly extracts collision avoidance hyperplanes from the point clouds within the visible range of the robot's sensor $R_\mathrm{sensor}$. 
Consequently, our approach substantially minimizes the computational and storage resources required for clustering obstacles.

 \begin{figure}[thpb]
  \centering
  \setlength{\belowcaptionskip}{-0.6cm}
\setlength{\abovecaptionskip}{0.3cm}
  \includegraphics[width =\hsize]{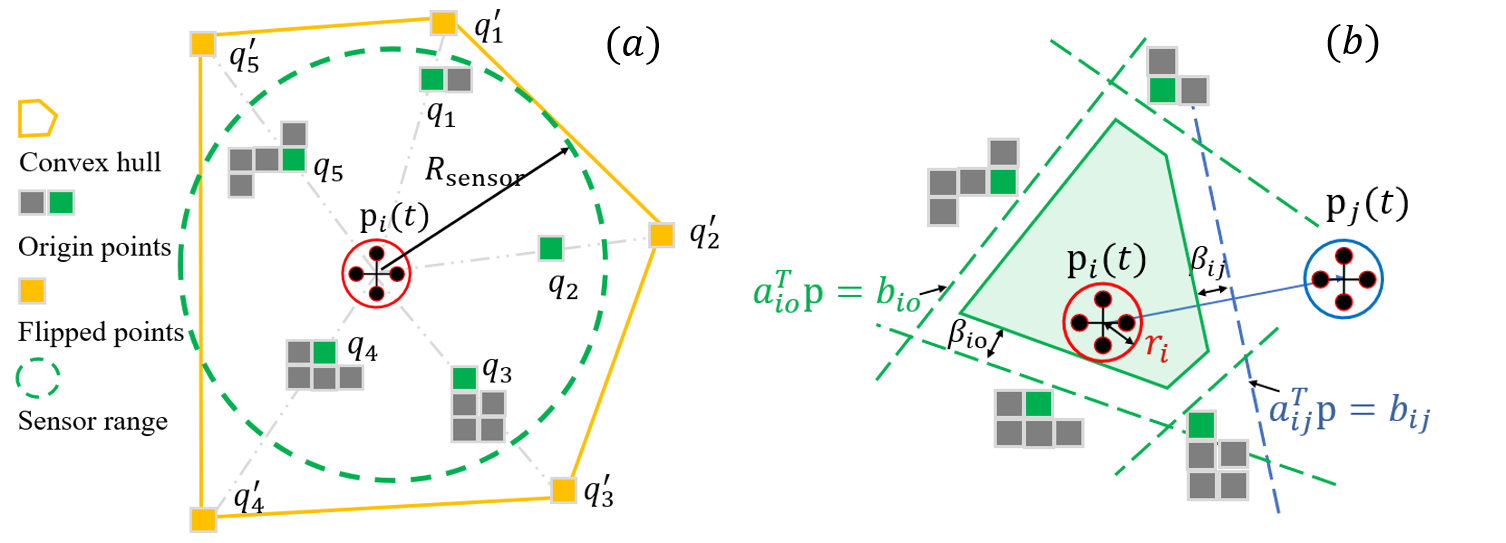}
\caption{Illustration of spatial decomposition and safe region construction in environments filled with a high density of obstacles, as represented by point clouds.
The original squares within sensor range $R_\mathrm{sensor}$ are projected outside of the circle with a one-to-one mapping. 
The robot is positioned in an obstacle-free location $\mathbf{p}_i$. 
By finding the convex hull of the mirrored points, i.e., $\mathbf{q}'_1,\dots,\mathbf{q}'_5$ (the yellow squares), we can target which squares that are closest to the robot, i.e., $\mathbf{q}_1,\dots,\mathbf{q}_5$ (the green squares). Subsequently, a safe convex region $\Bar{\mathcal{V}}_i$ colored in green can be obtained by separating hyperplane theorem.}
\label{fig:pt}
\end{figure}

Assuming robot $i$ is initially collision-free $\mathbf{p}_i \in  \mathcal{W}_{\mathrm{free}}$  surrounded by $m_r$ point clouds of obstacles inside the sensor range $R_\mathrm{sensor}$.
We use spherical mirroring operation mentioned in \cite{katz2005mesh} to map each $\mathbf{q}_o \in \mathcal{Q}$ to $\mathbf{q}_o'$ with $R_\mathrm{sensor}$:
\begin{equation}
\mathbf{q}_o' = F(\mathbf{q}_o) = \mathbf{q}_o - \mathbf{p}_i + 2(R_\mathrm{sensor} - \|\mathbf{q}_o - \mathbf{p}_i\|)\frac{\mathbf{q}_o - \mathbf{p}_i}{\Vert \mathbf{q}_o - \mathbf{p}_i \Vert},
\end{equation}
 where $\mathbf{q}_o'$ is the mirrored point, $\mathbf{q}_o$ is the original point, and $F(\mathbf{q}_o)$ is the spherical mapping function. As illustrated in Fig.~\ref{fig:pt} (a), the purpose of employing spherical mirroring is to reverse the positions of any points $\mathbf{q}_i$ that are located within the sphere along the corresponding ray. This mirroring operation effectively relocates the origin points of obstacles from the internal region of the sphere to the external region, creating a mirrored representation.
By using the QuickHull technique \cite{barber1996quickhull},
we can efficiently determine the transformation points required to construct a convex hull decided by a vertex vector
$\Omega_c= [\mathbf{q}_1',\dots, \mathbf{q}_{m_c}'] \in \mathbb{R}^{3\times m_c}$ 
from the mirrored points. 

Moreover, due to the monotonically decreasing nature of the function $\Vert F(\cdot) \Vert$, points that are closer to the robot are transformed further away. Since there is no point outside the convex hull $\Omega_c$, it implies that the area inside the corresponding origin points $\mathbf{q}_1,\dots,\mathbf{q}_{m_c}$, is obstacle-free. 
With the described spherical mirroring operation, the extraction of points that are necessary for generating robot-obstacle collision avoidance hyperplanes becomes a straightforward task for an unordered point cloud.
\subsection{ Safe Region Construction}

Based on above, $\mathbf{a}_{io}$ can be quickly calculated by solving the following low-dimensional quadratic programming (QP) problem:
\begin{equation}
 \begin{aligned}
 \label{svm}
 \min \quad &\mathbf{a}_{io}^\mathrm{T}\mathbf{a}_{io}\\
 \text{s.t.} \quad & (\mathbf{q}_o-\mathbf{p}_i)^\mathrm{T}\mathbf{a}_{io} \geq 1, \quad \forall o \in 1,\dots,m_c.
\end{aligned}
\end{equation}
We then shift the hyper-plane to be tight with the target. Thus, $b_{io} =\min{\mathbf{a}_{io}^\mathrm{T}\mathbf{q}_o}$. 
Solving a QP problem requires $O(N^3)$ time, where $N$ denotes the number of decision variables \cite{zhou2017fast}. These decision variables are linearly correlated with the number of obstacle points $m_c$ in our method. 
Consequently, the total computational complexity for each robot, influenced by the local density of obstacles, is $O(m_c^3)$. Leveraging advanced optimization utilities, such as CVXOPT, we can solve the problem in tens of milliseconds.

Moreover, we take into account the geometric dimensions of robots by employing a buffered term \cite{10236908}.	 
We introduce safety buffer variables $\beta_{ij} =  r_i \Vert \mathbf{a}_{ij}\Vert $
and $ \beta_{io} =  r_i\Vert \mathbf{a}_{io}\Vert $ to ensure the body of each robot within its corresponding buffered Voronoi cells (BVCs). 
It should be noted that these buffer variables can be further generalized to accommodate robots of varying dimensions along different axes. Then,  Eq.~\eqref{voronoi} is extended to the following definition. 
\begin{definition}
For a team of $n$ robots localized at $\mathbf{P}$ in an obstacle-free workspace $\mathcal{W}_\mathrm{free}$, the buffered Voronoi cell for each robot $i$ is: 
\begin{equation}
\begin{aligned}
\label{bvc}
\Bar{\mathcal{V}}_i = \{\mathbf{p} \in  \mathcal{W}_{\mathrm{free}}| & \mathbf{a}_{ij}^\mathrm{T}\mathbf{p}\leq b_{ij} -\beta_{ij},\forall j\neq i, i,j\in 1,\dots,n,\\
 & \mathbf{a}_{io}^\mathrm{T}\mathbf{p}\leq b_{io}-\beta_{io}, o\in 1,\dots,m_c \}.
\end{aligned} 
\end{equation}
\end{definition}
\begin{lemma}[Properties of BVC]
\label{lemma1}
    If $ \Vert \mathbf{p}_{i}(t) - \mathbf{p}_{j}(t) \Vert\geq r_i+r_j, i \neq j $
    and $\Vert \mathbf{p}_i(t) -\mathbf{q}_o \Vert \geq r_i$, $\forall i,j \in 1,\dots,n, o\in 1,\dots,m_c $ at time $t$, 
    we have 
    1) $\Bar{\mathcal{V}_i} \neq \emptyset$; 
    2)$\Bar{\mathcal{V}_i} \subset \mathcal{V}_i$; 
    3) $ \Vert \bar{\mathbf{p}}_i - \bar{\mathbf{p}}_j \Vert \geq r_i+r_j, \forall \bar{\mathbf{p}}_i \in \Bar{\mathcal{V}}_i, \bar{\mathbf{p}}_j \in \Bar{\mathcal{V}}_j \forall i \neq j$; 
    4) $\bar{\mathcal{V}}_i \cap \bar{\mathcal{V}}_j = \emptyset$;  
    5) $\Vert \bar{\mathbf{p}}_i - \mathbf{q}_o \Vert \geq r_i, \bar{\mathbf{p}}_i \in \Bar{\mathcal{V}}_i, \mathbf{q}_o \in \Omega$;
    6) $\mathbf{q}_o \notin \bar{\mathcal{V}}_i$. 
\end{lemma}

\emph{Proof:}
According to Section II-A in \cite{zhou2017fast}, 1) - 4) have been proved.
5) According to Eq.~\eqref{bvc} and Eq.~\eqref{svm}, $\mathbf{a}_{io}^\mathrm{T}\bar{\mathbf{p}}_i\leq b_{io}-\beta_{io}$,
$\mathbf{a}_{oi}^\mathrm{T}\mathbf{q}_o\leq b_{oi}$.
Since $\mathbf{a}_{io} = -\mathbf{a}_{oi}$ and $b_{io} = -b_{oi}$, by adding them, we get $\mathbf{a}_{io}^\mathrm{T}(\bar{\mathbf{p}}_i - \mathbf{q}_o)  \leq -\beta_{io}$. Due to $ \beta_{io} = r_i \Vert \mathbf{a}_{io}\Vert \in \mathbb{R}^+$, we have $\Vert \mathbf{a}_{io}^\mathrm{T}(\bar{\mathbf{p}}_i - \mathbf{q}_o) \Vert \geq   r_i \Vert \mathbf{a}_{io}\Vert$. Since $ \Vert \mathbf{a}_{io}\Vert \Vert \bar{\mathbf{p}}_i - \mathbf{q}_o \Vert \geq \Vert \mathbf{a}_{io}^\mathrm{T}(\bar{\mathbf{p}}_i - \mathbf{q}_o) \Vert  $
Therefore, $\Vert\bar{\mathbf{p}}_i - \mathbf{q}_o \Vert \geq r_i$.
6) If $\mathbf{q}_o\in \Bar{\mathcal{V}}_i$, then Eq.~\eqref{bvc} should be satisfied, i.e., $\mathbf{a}_{io}^\mathrm{T}\mathbf{q}_o \leq b_{io}-\beta_{io}$. Rewritten this, we have $\mathbf{a}_{io}^\mathrm{T}\mathbf{q}_o + r_i \Vert \mathbf{a}_{io} \Vert \leq b_{io}$.  According to 5), $\mathbf{a}_{oi}^\mathrm{T}\mathbf{q}_o\leq b_{oi}$, i.e., $\mathbf{a}_{io}^\mathrm{T}\mathbf{q}_o\geq b_{io}$, which  contradicts to the definition of $\Bar{\mathcal{V}}_i$.
\rightline{$\Box$}

By combining separating hyperplane theorem and sphere flipping transformation, our method can efficiently extract a collision-free region using only raw measurement points, as shown in Fig.~\ref{fig:pt} (b).
Consequently, each robot is capable of independently calculating its operational domain $\mathcal{V}_i$  based on the relative positioning of its peers in a decentralized fashion at each time step.  
\begin{theorem}
If a team of robot is initialized at collision-free configuration, i.e., $\Vert  \mathbf{p}_i(0)-\mathbf{p}_j(0)\Vert \geq r_i+r_j, \forall i\neq j , \Vert\mathbf{p}_i(0) - \mathbf{q}_o\Vert \geq r_i$, by using Eq.~\eqref{eq: u}, then for $\forall t > 0$, we have $\Vert\mathbf{p}_i(t)-\mathbf{p}_j(t) \Vert \geq r_i+r_j, \forall i\neq j, \Vert\mathbf{p}_i(t) - \mathbf{q}_o\Vert \geq r_i $.
\end{theorem}

\emph{Proof:}
Given that the robot is initially collision-free, according to Lemma~\ref{lemma1}, safe $\Bar{\mathcal{V}}_i$ can be calculated considering obstacle and other robots information using Eq.~\eqref{bvc}.
Since the policy Eq.~\eqref{eq: u} ensures the robot's position  $\mathbf{p}_i$ is updated inside its corresponding Voronoi cell, i.e., $\mathbf{C}_{\Bar{\mathcal{V}}_i} \in \Bar{\mathcal{V}}_i$, by employing mathematical induction, it can be easily proved that the robots' movements will remain confined to a sequence of secure convex regions for all future time \cite{zhou2017fast}. \rightline{$\Box$}

Therefore, constructing these safe convex regions, the multi-robot system is designed to strictly prevent duplicated task execution and collisions. 

\subsection{Deadlock-aware Guided Map}




Despite its efficiency
 in convex environments, the control policy in Eq.~\eqref{eq: u} may let the robot get stuck in many realistic scenarios. These challenges arise primarily due to the complex, non-convex arrangements of obstacles.
To address this, we introduce a real-time constructed guided map that is dynamically adjusted in response to environmental changes, thereby enabling the robot to avoid pitfalls and enhancing both the efficiency of coverage and the system's adaptability.

In the proposed method, we consider the density map $\phi$ in Eq.~\eqref{eq: cost}, which characterizes the distribution of information.  Each point on the grid map is assigned a corresponding value. As the robot moves, it employs local sensing to identify a point $\mathbf{p}_g$ within its sensor range $R_\mathrm{sensor}$ that represents a local maximum in the information distribution, which then becomes the robot's navigation goal. 
To facilitate reaching this local objective, 
we define the navigation function (NF)  $\mathcal{M}_\text{go}: \mathcal{W}_{\mathrm{free}} \rightarrow \mathbb{R}^{+}$ as a continuous function that approximates the minimum length of a collision-free path from any point $\mathbf{q} \in \mathcal{W}_{\text{free}}$ to $\mathbf{p}_g$, i.e., cost-to-goal function.
\begin{lemma}[Properties of NF \cite{lavalle2006planning}]
\label{lemma2}
1) $\mathcal{M}_\text{go}(\mathbf{p}_g) = 0$, and
$\mathcal{M}_\text{go}(\mathbf{p}) = \infty$ iff no point in $\mathcal{W}$ is reachable from $\mathbf{p}$, e.g., obstacle position $\mathbf{q}_o, o\in 1,\dots,m_c$; 
2) 
For every reachable position $\mathbf{p}(t)$ at time $t$, executing an action yields a subsequent position $\mathbf{p}(t+\Delta t)$ that satisfies $\mathcal{M}_\text{go}(\mathbf{p}(t+\Delta t)) < \mathcal{M}_\text{go}(\mathbf{p}(t))$.
\end{lemma}

From Lemma~\ref{lemma2}, it can be observed that the NF exhibits no local minima other than at the goal, ensuring a cycle-free execution of actions that inevitably leads to the goal position.
According to \cite{lavalle2006planning}, the NF determined by Dijkstra’s algorithm working backward from the goal yields an optimality. Here, to balance the efficiency and optimality, we utilize A$^*$ algorithm to compute cost-to-go values for each grid on a voxel-based map representation. 
Consequently, based on $\mathcal{M}_\text{go}$, the map $\phi$ in the definition of $\mathcal{H}(\mathbf{P})$ in the Eq.~\eqref{eq: cost} is appropriately modified as follows:
\begin{equation}
\begin{aligned}
\label{eq:phi}
\phi = e^{-\gamma \mathcal{M}_\text{go}(\mathbf{p})}, \forall \mathbf{p} \in \mathcal{W}_\mathrm{free},
\end{aligned} 
\end{equation}
where gain $\gamma \geq 0$ serves as a design parameter that influences the magnitude of the value.
By integrating centroid Voronoi tessellation with $\phi$ and computing $\mathbf{C}_{\Bar{\mathcal{V}}_i}$, the regions are shifted to align with higher values in the density map, thereby providing feasible heading directions for coverage.   
The negative gradient of $\mathcal{M}_\text{go}$ always effectively shortens the shortest path to the destination. 

\begin{lemma}[Convergence \cite{pierson2017distributed}]
Assuming there is a finite set of $\mathcal{W}$, by adopting controller in Eq.~\eqref{eq: u}, the location of each sensor will decrease its configuration cost and asymptotically converge to a static location. 
\end{lemma}


 
This approach synergizes local sensing with global information distribution, resulting in more optimal coverage outcomes. 
Finally, a sequence of coverage path during the time
interval reflecting the optimized location of deployment for each robot can be obtained.

\section{Simulations}
This section analyzes the applicability of the proposed method across diverse and unstructured settings, including real forests and indoor office datasets, and evaluates the performance of computational time and collision distance. 
Besides, benchmarking our method against established coverage control protocols can help quantify the improvements in task efficiency.
The accompanying video provides a clear visual aid to understand our method's mechanics and advantages.\footnote{\href{https://youtu.be/wOunvjGHhBQ}{https://youtu.be/wOunvjGHhBQ}}
The simulations are conducted on a laptop equipped with an Intel i7-9700 CPU and 16GB of RAM. A team of robots is simulated through multiprocessing. For real-world scenario validation, We utilized the Robot Operating System (ROS) with C++ and Python programming languages. MATLAB is employed for comparative analysis. 
Velocity constraints are established with a maximum limit of $u_\text{max}=2.5$m/s. The time interval for each replanning is $\Delta t =0.1$s. The obstacles are inflated by a robot radius, $r_i$= 0.25m. The grid map resolution is also set to 0.25m for all axes. The sensor's radius is restricted to a range of 15m. If $\Vert \mathbf{u} \Vert <0.01$m/s, the robot is considered to have converged.

\subsection{High-fidelity Validation}
Unlike other coverage methods, like Obstacle-aware Voronoi Cells (OAVC) \cite{pierson2017distributed,abdulghafoor2021distributed}, which are restricted to hyperspherical object models, or the method in \cite{wang2023fast}, which presumes obstacles to be convex polytopes, our approach allows for a broader range of obstacle shapes with various obstacle densities. 

\begin{figure}[thpb]
  \centering
\setlength{\belowcaptionskip}{-0.3cm}
\setlength{\abovecaptionskip}{0.2cm}
  \includegraphics[width =\hsize]{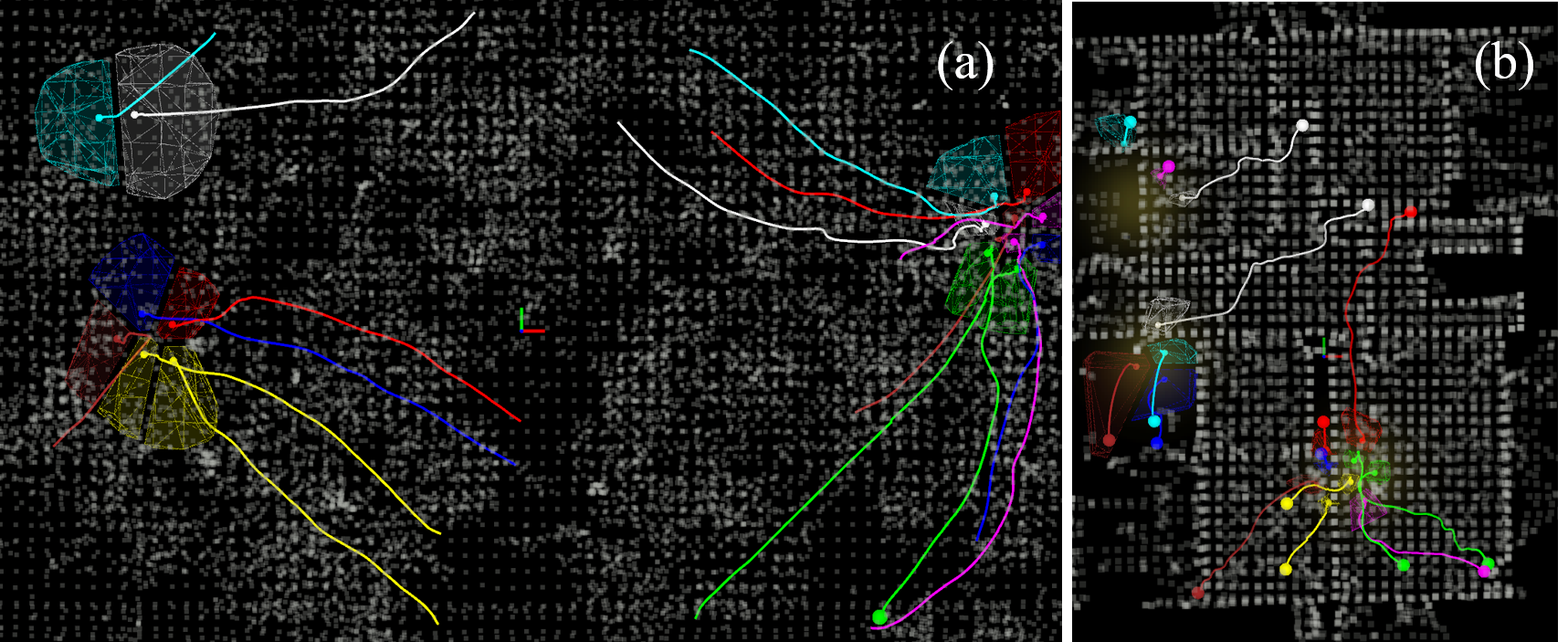}
  \caption{Coverage of ToI in two 3D large-scale scenarios by 16 robots. (a) Cluttered and thin structural forest environment. (b) Narrow indoor office environment.}
  \label{fig:forest}
\end{figure}

We execute tests utilizing high-fidelity point cloud data, sourced from forested areas by the University of Hong Kong and facilitated by the MARSIM simulator \cite{kong2023marsim}. 
This dataset is characterized by extensive cluttered and thin structural elements. 
We expanded the scope of the forest environment to large-scale dimensions of 140m $\times$ 140m $\times$ 10m, and the coverage result is depicted in Fig.~\ref{fig:forest}.
We also use data from indoor office settings, courtesy of the Technical University of Munich \cite{ikehata2015structured} with dimensions of 80m $\times$ 80m $\times$ 5m. This environment
featured narrow passageways and multiple layers, and the coverage result is depicted in Fig.~\ref{fig:forest}.


\begin{figure}[!t]
\centering
\setlength{\belowcaptionskip}{-0.6cm}
\setlength{\abovecaptionskip}{0.3cm}
\hspace{-0.4cm}
\vspace{-1.5pt}
\subfloat{
\includegraphics[width=1.7in]{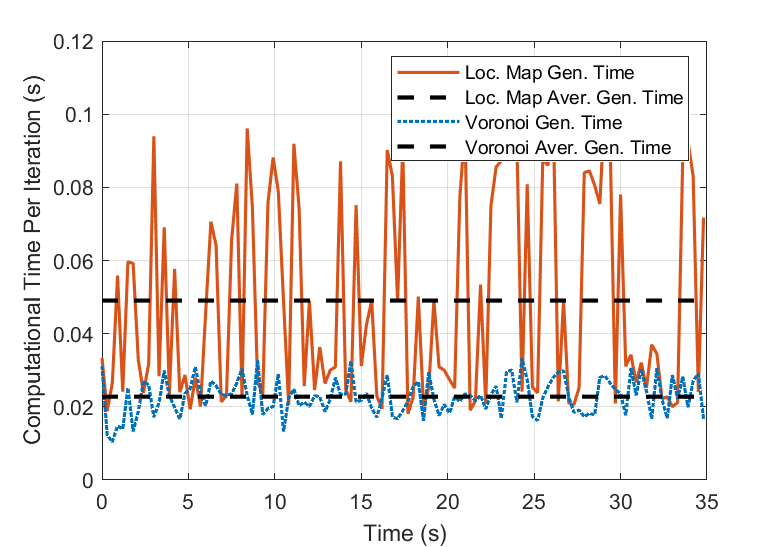}
\label{fig:data2_com}}s
\hfil
\hspace{-0.71cm}
\vspace{-1.5pt}
\subfloat{
\includegraphics[width=1.7in]{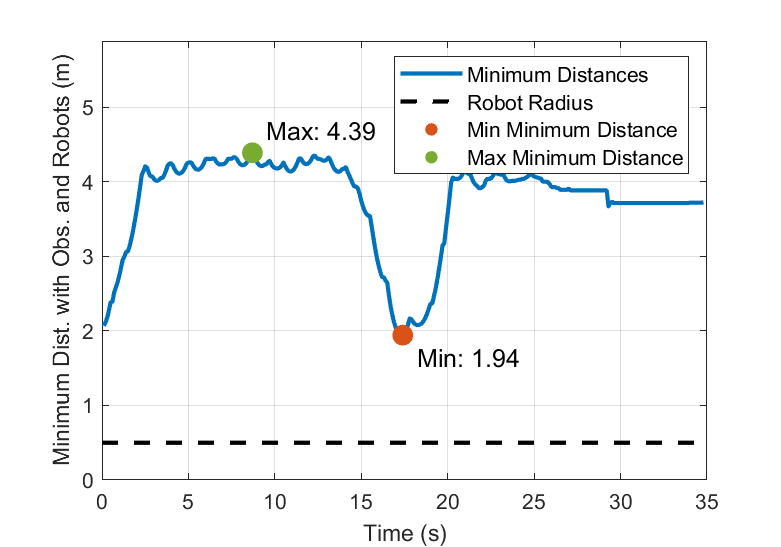}
\label{fig:data2_min}}
\hfil
\caption{One trial performance in terms of computational time and minimum distance during coverage by 16 robots.}
\label{fig:data1}
\end{figure}

We tested the above mentioned two scenarios with randomized initialization and ToI. 
We evaluate the two most time-consuming components, local map and Voronoi cell generation, and consider both neighboring robots and obstacles for evaluating collision avoidance performance.
One trial test in the forest is shown in Fig.~\ref{fig:forest} (a) and Fig.~\ref{fig:data1}. The time required for generating a local map depends on the complexity of computing a feasible path. This complexity increases with a high density of obstacles, where conditions such as non-convex obstacles can potentially trap the robots. Despite these challenges, our proposed method can still enable robots to create a secure area in real-time while ensuring safety margins (1.95m) that substantially exceed the robot's radius (0.25m).

\begin{figure}[!t]
\centering
\setlength{\belowcaptionskip}{-0.6cm}
\setlength{\abovecaptionskip}{0.3cm}
\hspace{-0.4cm}
\vspace{-1.5pt}
\subfloat{
\includegraphics[width=1.7in]{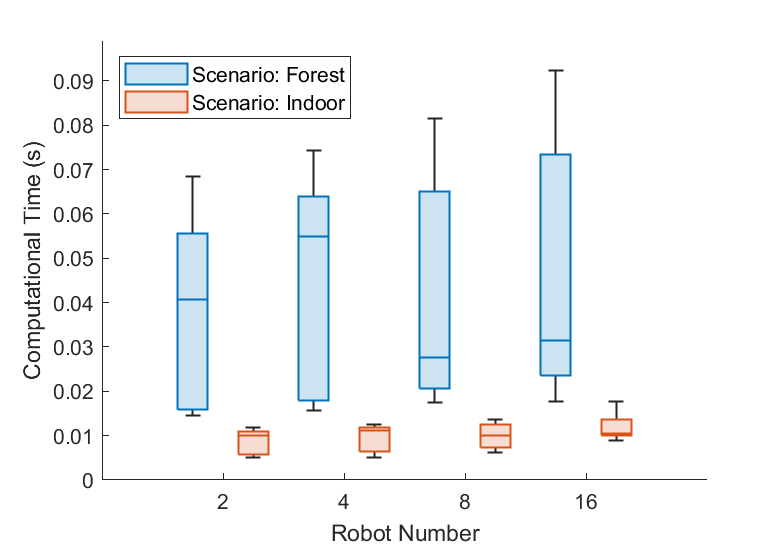}
\label{fig:data1_com}}
\hfil
\hspace{-0.71cm}
\vspace{-1.5pt}
\subfloat{
\includegraphics[width=1.7in]{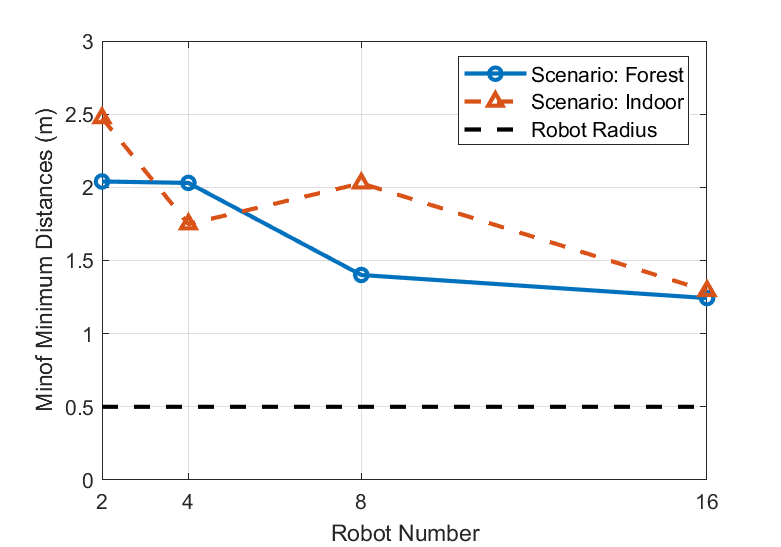}
\label{fig:data1_min}}
\hfil
\caption{The variation between computational time and minimum distance with the increasing number of robots in two scenarios with different obstacle densities.}
\label{fig:data2}
\end{figure}

We also tested with varying robot numbers (n=2, 4, 8, 16), each configuration undergoing 10 trials, as depicted in Fig.~\ref{fig:data2}. Due to the decentralized nature of our algorithm, computational time remains largely invariant as the number of robots scales up. Additionally, in more expansive environments—nearly double in obstacle density, the algorithm still satisfies real-time processing criteria. The incidence of collisions remains zero, even when covering obstacle-dense areas, and there are only negligible fluctuations in the minimum distance
as the number of robots increases.
This demonstrates both scalability in larger contexts and robustness in intricate environments of the proposed method.

\begin{figure}[!t]
\centering
\setlength{\belowcaptionskip}{-0.6cm}
\setlength{\abovecaptionskip}{0.3cm}
\hspace{-0.4cm}
\vspace{-1.5pt}
\subfloat{
\includegraphics[width=1.75in]{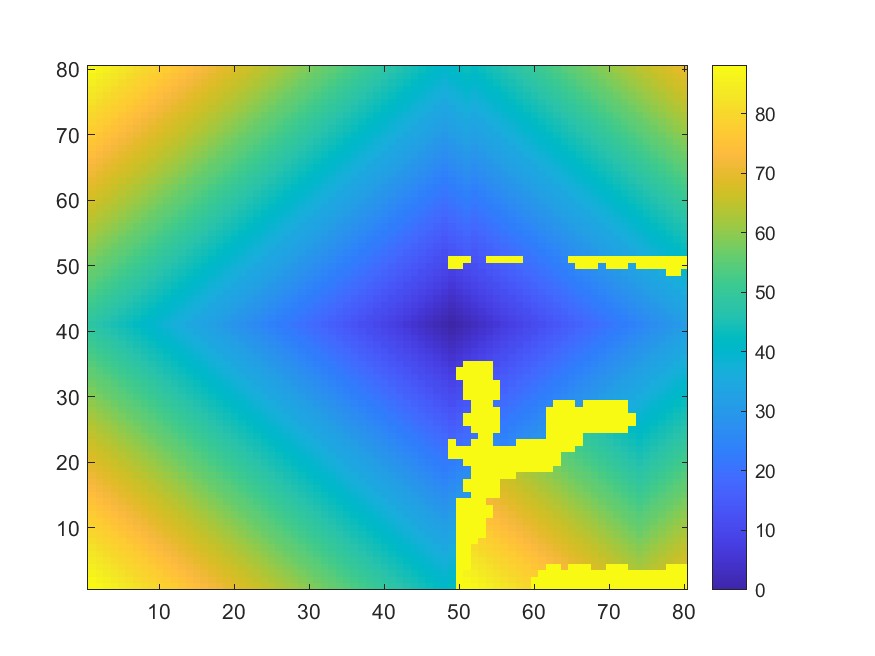}
\label{xx}}
\hfil
\hspace{-0.71cm}
\vspace{-1.5pt}
\subfloat{
\includegraphics[width=1.75in]{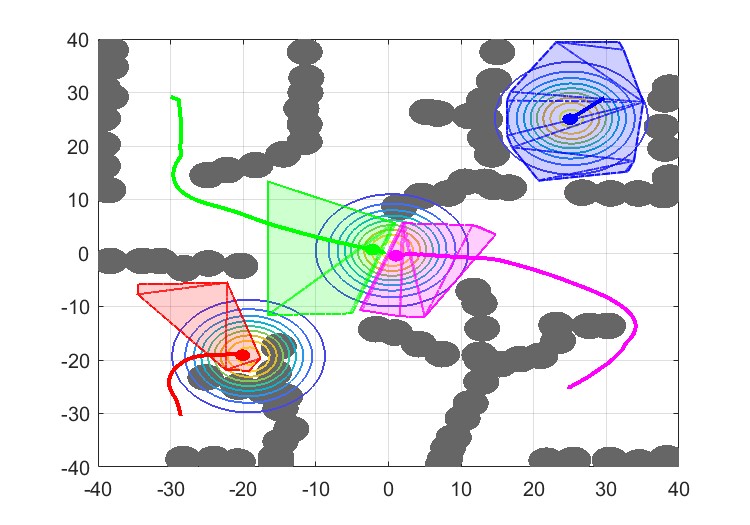}
\label{xxx}}
\hfil
\caption{Coverage in a cluttered environment using the proposed method and Deadlock-aware guided map of the pink robot (t=12.10s). The black points represent obstacles, and the colored contour lines represent the ToI with three peaks.}
\label{fig:map}
\end{figure}

\begin{figure}[!t]
\centering
\setlength{\belowcaptionskip}{-0.6cm}
\setlength{\abovecaptionskip}{0.3cm}
\hspace{-0.4cm}
\vspace{-1.5pt}
\subfloat[OAVC \cite{abdulghafoor2021distributed}]{
\includegraphics[width=1.75in]{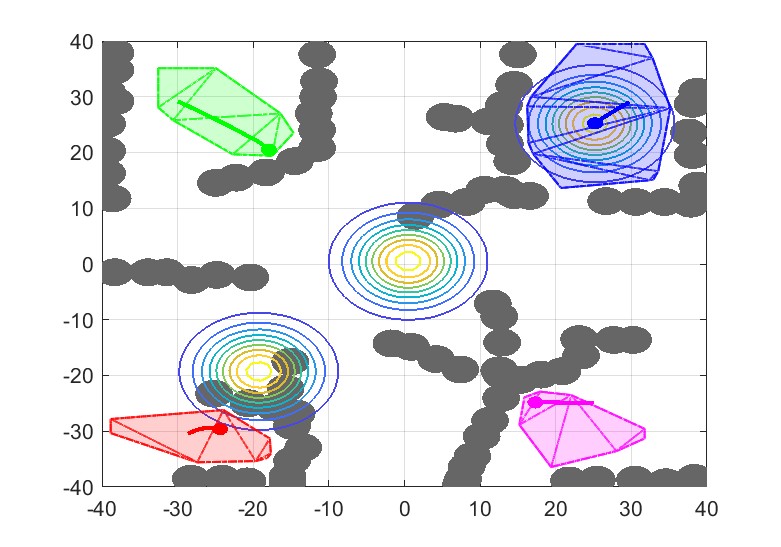}
\label{fig:comoavc}}
\hfil
\hspace{-0.71cm}
\vspace{-1.5pt}
\subfloat[Adaptive method \cite{bai2022adaptive}]{
\includegraphics[width=1.75in]{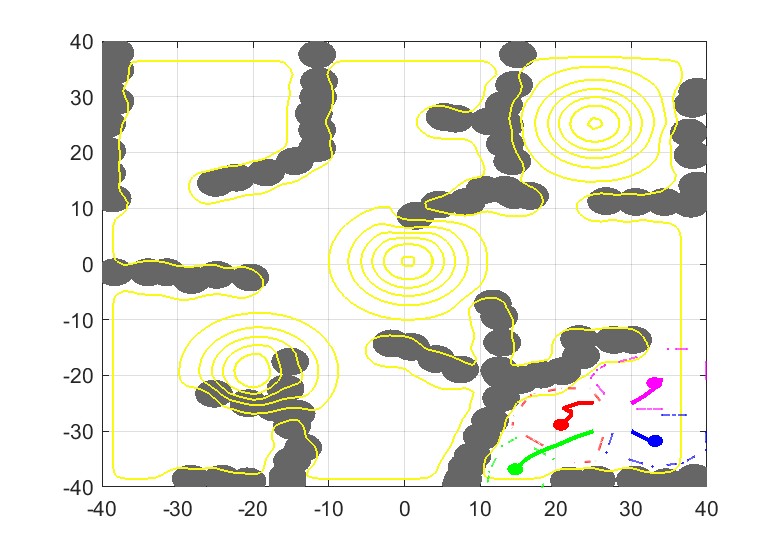}
\label{fig:compare2}}
\hfil
\caption{Compared coverage path obtained by OAVC and adaptive method when convergence. (a) The colored contour lines represent the ToI computed by a Gaussian density map. (b) The yellow contour lines represent the density map, combining both attractive (ToI coverage) and repulsive fields.
}
\label{fig:compare}
\end{figure}


\subsection{Comparisons of Coverage Performance}
We undertake comparative evaluations with other Voronoi-based coverage methods: 1) OAVC method in \cite{abdulghafoor2021distributed}, formulates secure areas by computing tangential boundary lines to circular obstacles and adheres to the gradient of target information; 2) Adaptive coverage method in \cite{bai2022adaptive},  utilizes a repulsive density function concerning detected obstacles, ensuring that the newly computed centroid is positioned away from the obstacle.
To be fair, the obstacles are modeled as the same shape in \cite{abdulghafoor2021distributed}. We evaluate these methods over 30 trials, each with randomly initialized starting positions within an 80m $\times$ 80m $\times$ 16m workspace. The ToI is defined by a Gaussian distribution featuring three peaks. The performance metrics are as follows: 1) Coverage ratio: The ratio of detected peaks to the total number of peaks; 
2) Success rate: The fraction of trials where robots successfully converge to the ToI without experiencing deadlock or collisions within a limited time (30s) or until all peaks are detected; 3) Average time: Task execution time, considering only successful trials. The results are shown in Tab.~\ref{tab:per}.

As shown in Fig.~\ref{fig:map}, utilizing the proposed method, even in highly irregular environments, all robots can maintain safe distances between each other and prevent both local minima and potential collisions.
In contrast, in Fig.~\ref{fig:compare} (a),
the OAVC method directs the robots toward areas of high density without accounting for the obstacle effect, thereby resulting in getting stuck in corners.
In Fig.~\ref{fig:compare} (b), when multiple repulsive fields and attractive fields (target coverage) interact with each other, the adaptive method in \cite{bai2022adaptive} predisposes robots to entrapment. Its constrained avoidance space further exacerbates this issue, as it can reduce the navigable space and lead a group of robots to be trapped when moving collectively, thereby increasing the risk of failure.


\begin{table}[htbp]
\centering
\caption{Comparisons of coverage performance}
\begin{tabular}{@{}ccccc@{}}
\toprule
Alg. & Sensor & Cover. Ratio  & Succ. Rate & Aver. Time \\ 
\midrule
Ours  & Local & 27/30 & 96.67\% & 15.48s\\ 
OAVC \cite{abdulghafoor2021distributed}  & Global & 22/30 & 46.67\% & 17.14s\\ 
Adaptive \cite{bai2022adaptive}   & Local& 13/30 & 10.00\% & 13.30s\\
\bottomrule
\end{tabular}
\label{tab:per}
\end{table}



\section{Conclusions}

In this work, our proposed method, which incorporates collision avoidance without entrapment, is effectively applicable across a range of unstructured settings. The approach leverages spatial decomposition and sphere flipping transformation to construct a safe Voronoi region in complex environments with considerable point cloud data. The NF is integrated to steer the robot away from pitfalls while simultaneously directing it towards the ToI.
Comparisons with others highlight that our approach performs extremely well, with high success ratios and guaranteed safety. Our method also extends the traditional coverage control method, move-to-centroid policy, into more realistic scenarios.
Future work includes testing our algorithm on physical robot swarms to address hardware-specific challenges and to further validate its real-world applicability.


%


\bibliographystyle{IEEEtran}
\bibliography{refs}

\end{document}